# A CNN-RNN Framework for Crop Yield Prediction

Saeed Khaki[1*], Lizhi Wang[1] and Sotirios V. Archontoulis[2]

[1] Industrial and Manufacturing Systems Engineering Department, Iowa State University, Ames, IA, United States,
[2] Department of Agronomy, Iowa State University, Ames, IA, United States

Crop yield prediction is extremely challenging due to its dependence on multiple factors such as crop genotype, environmental factors, management practices, and their interactions. This paper presents a deep learning framework using convolutional neural networks (CNNs) and recurrent neural networks (RNNs) for crop yield prediction based on environmental data and management practices. The proposed CNN-RNN model, along with other popular methods such as random forest (RF), deep fully connected neural networks (DFNN), and LASSO, was used to forecast corn and soybean yield across the entire Corn Belt (including 13 states) in the United States for years 2016, 2017, and 2018 using historical data. The new model achieved a root-mean-square-error (RMSE) 9% and 8% of their respective average yields, substantially outperforming all other methods that were tested. The CNN-RNN has three salient features that make it a potentially useful method for other crop yield prediction studies. (1) The CNN-RNN model was designed to capture the time dependencies of environmental factors and the genetic improvement of seeds over time without having their genotype information. (2) The model demonstrated the capability to generalize the yield prediction to untested environments without significant drop in the prediction accuracy. (3) Coupled with the backpropagation method, the model could reveal the extent to which weather conditions, accuracy of weather predictions, soil conditions, and management practices were able to explain the variation in the crop yields.

# INTRODUCTION

Crop yield is affected by many factors such as crop genotype, environment, and management practices. Crop genotype has improved significantly over years by seed companies. Environments, changing spatially and temporally, have huge effects on year-to-year and location-to-location variations in crop yield (Horie et al., 1992). Under such circumstances, accurate yield prediction is very beneficial to global food production. Timely import and export decisions can be made based on accurate predictions. Farmers can utilize the yield prediction to make knowledgeable management and financial decisions. Performances of new hybrids can be predicted in new and untested locations (Khaki and Wang, 2019). However, successful crop yield prediction is very difficult due to many complex factors. For example, genotype and environmental factors often have interactions with each other, which makes the yield prediction more challenging. Environmental factors such as weather components often have complex nonlinear effects, which are difficult to estimate accurately.

Many studies have used machine learning techniques such as regression tree, random forest, multivariate regression, association rule mining, and artificial neural networks for crop yield prediction. Machine learning models treat the output, crop yield, as an implicit function of the input variables such as weather components and soil conditions, which could be a very complex and nonlinear function. Jeong et al. (2016) applied random forest and multiple linear regression for yield prediction of wheat, maize, and potato. They found that random forest was highly capable of predicting crop yields and outperformed multiple linear regression. Fukuda et al. (2013) also used random forest for predicting mango fruit yields in response to water supply under different irrigation regimes, and found that random forest was applicable for mango yield prediction with a specific focus on water management. Liu et al. (2001) applied artificial neural networks to approximate a nonlinear function to relate the corn yield to input variables such as weather, soil, and management practices. Ransom et al. (2019) evaluated machine learning methods for corn nitrogen recommendation tools using soil and weather information. Drummond et al. (2003) investigated stepwise multiple linear regression, projection pursuit regression, and artificial neural networks to predict the grain yield based on the soil properties. Shahhosseini et al. (2019) predicted corn yield and nitrate loss using machine learning algorithms such as random forest and multiple linear regression. Awad (2019) designed a mathematical optimization model to predict potato yield using the biomass calculated by the model. Jiang et al. (2004) applied artificial neural network and multiple linear regression for estimating winter wheat yields based on the remotely sensed and climate data, and found that artificial neural network model outperformed the multiple linear regression. Prasad et al. (2006) used piecewise linear regression method with breakpoint to predict corn and soybean yields based on remote sensing data and other surface parameters. Romero et al. (2013) applied several machine learning methods such as decision tree and association rule mining for the classification of yield components of durum wheat and showed that association rule mining method obtained the best performance across all locations.

This paper presents a deep learning framework that takes advantage of the state-of-the-art modeling and solution techniques to predict crop yield based on environmental data and management practices. Deep learning methods belong to the class of representation learning methods with multiple levels of representation, each having nonlinear modules to transform the representation at the current level (starting with the raw input) to a slightly more abstract level (LeCun et al., 2015). Deep neural networks also provide a universal approximation framework, which means that regardless of what function we want to learn, deep neural networks can be used to represent such function (Hornik et al., 1989; Goodfellow et al., 2016). Deep learning methods do not require handcrafted features and they learn the features from data, which contribute to the higher accuracy of results (LeCun et al., 2015).

In comparison with the aforementioned artificial neural network models in the literature, which had a single hidden layer, deep learning methods with multiple hidden layers tend to perform better. However, deeper models are more difficult to train and require more advanced hardware and optimization techniques (Goodfellow et al., 2016). For example, the loss function of the deep neural networks is extremely high dimensional and non-convex, which makes the optimization of such function more difficult due to having many local optima and saddle points (Goodfellow et al., 2016). Deeper networks may also have the vanishing gradient problem, which can be alleviated by using residual shortcut connections or multiple auxiliary heads (loss functions) for the network (Bengio et al., 1994; Szegedy et al., 2015; Goodfellow et al., 2016; He et al., 2016). Some other techniques have also been developed to improve performance of deep learning models such as batch normalization (Ioffe and Szegedy, 2015), dropout (Srivastava et al., 2014), and stochastic gradient descent (SGD) (Goodfellow et al., 2016).

Our proposed hybrid CNN-RNN model consists of convolutional neural networks (CNNs) and recurrent neural networks (RNNs). CNNs process data with multiple arrays format such as one-dimensional data (signals and sequences), two-dimensional data (images), and three-dimensional data (videos). A CNN model is usually composed of multiple convolutional and pooling layers followed by few fully connected (FC) layers. CNNs have some design parameters, including the number of filters, filter size, type of padding, and stride. A filter is a matrix of weights with which we convolve the input data. Padding is the process of adding zeroes to the input to preserve the dimension of the input space. The stride is the amount by which the filter is moved. RNNs are used for tasks involving sequential data to capture their time dependencies (LeCun et al., 2015; Sherstinsky, 2018). RNNs keep the history of all the past elements of a sequence in their hidden units called a state vector and use this information as they process input sequence one element at a time (LeCun et al., 2015). RNNs are very powerful models for sequence modeling, but training them has proved to be very challenging due to vanishing and exploding gradient problems (Bengio et al., 1994). To solve this problem, RNNs are improved by long short-term memory (LSTM) cells, which are carefully designed recurrent neurons giving superior performance in a wide range of sequence modeling applications (Hochreiter and Schmidhuber, 1997; Pham et al., 2014; Sherstinsky, 2018). LSTM cells use a special unit called the memory cell to remember inputs for a long time and prevent the vanishing gradient problem (LeCun et al., 2015).

More recently, deep learning methods have been applied for the crop yield prediction. Khaki and Wang (2019) designed a deep neural network model to predict corn yield across 2,247 locations between 2008 and 2016. Their model was found to outperform other methods such as Lasso, shallow neural networks, and regression tree. You et al. (2017) applied CNNs and RNNs to predict soybean yield based on a sequence of remotely sensed images. Kim et al. (2019) developed a deep neural network model for crop yield prediction using optimized input variables from satellite products and meteorological datasets between 2006 and 2015. Wang et al. (2018) designed a deep learning framework to predict soybean crop yields in Argentina and they also achieved satisfactory results with a transfer learning approach to predict Brazil soybean harvests with a smaller amount of data. Yang et al. (2019) investigated the ability of CNN to estimate rice grain yield using remotely sensed images and found that CNN model provided robust yield forecast throughout the ripening stage. Khaki and Khalilzadeh (2019) used deep CNNs to predict corn yield loss across 1,560 locations in the United States and Canada.

The remainder of this paper is organized as follows. The section *Data* introduces the data used in this paper. The section *Methodology* describes our proposed model for crop yield prediction. The section *Design of Experiments* provides implementation details of the models used in this research. The section *Results* presents the results. The section *Analysis* provides the analysis performed based on the proposed model. Finally, the section *Conclusion* concludes the paper.

## DATA

The data analyzed in this paper included four sets: yield performance, management, weather, and soil; no genotype data was found to be publicly available to complement these four datasets.

- The yield performance dataset contained observed average yield for corn and soybean between 1980 and 2018 across 1,176 counties for corn and 1,115 counties for soybean within 13 states of the Corn Belt: Indiana, Illinois, Iowa, Minnesota, Missouri, Nebraska, Kansas, North Dakota, South Dakota, Ohio, Kentucky, Michigan, and Wisconsin, in which corn and soybean are the dominant crops. **Figure S1** in the **Supplementary Material** shows the map of the Corn Belt in the United States.
- The management data included the weekly cumulative percentage of planted fields within each state, starting from April of each year. The yield performance and management data were acquired from National Agricultural Statistics Service of the United States (USDA-NASS, 2019).
- The weather data included daily record of six weather variables, namely precipitation, solar radiation, snow water equivalent, maximum temperature, minimum temperature, and vapor pressure. The weather data was acquired from Daymet (Thornton et al., 2018). The spatial resolution of the weather data was 1 km$^2$.
- The soil data included wet soil bulk density, dry bulk density, clay percentage, upper limit of plant available water content, lower limit of plant available water content, hydraulic conductivity, organic matter percentage, pH, sand percentage, and saturated volumetric water content variables measured at depths 0–5, 5–10, 10–15, 15–30, 30–45, 45–60, 60–80, 80–100, and 100–120 cm. Four soil variables were only recorded at the soil surface, which included field slope in percent, national commodity crop productivity index for corn, average national commodity crop productivity index for all crops, and crop root zone depth. The soil data was acquired from Gridded Soil Survey Geographic Database for the United States (gSSURGO, 2019). The spatial resolution of the soil data was 1 km$^2$.

We selected multiple weather and soil samples from each county based on the grid map approach and took the average of these samples to get representative samples for both weather and soil. The soil data had 6.7% missing values for some locations, which we imputed using the mean of the same soil variable of other counties. The management data had 6.3% missing values for some locations, which we imputed using the mean of the same management variable of other counties at the same year. We tried other imputation techniques such as median and most frequent and found that the mean approach led to the most accurate results.

The weather data did not have any missing values, but we found the daily data to be more granular than necessary to reveal the essential information. As a result, we took the weekly average and achieved a 365:52 ratio of dimension reduction. Such pre-processing of the weather data substantially reduced the number of trainable parameters of the first layer of the neural network model. **Table S1** in the **Supplementary Material** shows the summary statistics of the data corn and soybean.

## METHODOLOGY

The model that we propose for crop yield prediction is a hybrid one, which combines CNNs, fully connected layers, and RNNs, as illustrated in **Figure 2**. Details of this model are provided as follows.

### W-CNN and S-CNN

The W-CNN and S-CNN models were designed to capture the linear and nonlinear effects of the weather and soil data, respectively. The W-CNN model used one-dimensional convolution to capture the temporal dependencies of weather data, whereas the S-CNN model used one-dimensional convolution to capture the spatial dependencies of soil data measured at different depths underground. Similar convolutional models have been widely used in various application domains and found to be effective in improving prediction accuracy (Ince et al., 2016; Borovykh et al., 2017; Kiranyaz et al., 2019).

### FC

A fully connected layer (FC) was used to combine the high-level features of weather components and soil conditions extracted by W-CNN and S-CNN, which also reduced the dimension of the output of the CNN models.

### RNN

The RNN model was designed to capture the time dependencies of crop yield over a number of years. The use of the RNN model was motivated by two contrasting observations. On the one hand, both corn yield and soybean yield have demonstrated an increasing trend over the past four decades, as shown in **Figure 1**, which could be largely attributed to continuous improvement in genetics and management practices, thanks to significant research and development investment in breeding and farming techniques. On the other hand, genotype data was not publicly available for this prediction study. Therefore, the effect of genotype must be indirectly reflected in the model using available data. RNN is a type of artificial neural network where temporal dependencies of nodes are reflected with a directed graph. So we designed a special RNN model to capture the temporal dynamic behavior of crop yield as a result of genetic improvement. These RNNs were enhanced with LSTM cells, which are carefully designed recurrent neurons to capture dependencies of input with time. Compared with other time series models, LSTM networks do not need to specify the nonlinear functions to be estimated, and they have demonstrated superior performance in a wide range of sequence modelling applications (Hochreiter and Schmidhuber, 1997; Pham et al., 2014; Sherstinsky, 2018).

The RNN model consisted of $k$ LSTM cells, which predicted crop yield of a county for year $t$ using information from years $t - k$ to $t$. Input to the cell includes average yield (over all counties in the same year) data, management data, and output of the FC layer, which extracted important features processed by the W-CNN and S-CNN models using the weather and soil data. The only exception is that the S-CNN and FC models were specifically designed to pass the soil data measured at the soil surface directly to the LSTM cells. Although the soil data is generally static, the subscript of soil data in **Figure 2** allows the possibility of changing soil conditions over time. The use of historical average yield data as part of the input allows the RNN model to predict crop yield using historical trend of crop yield even without weather or soil data. Suppose year $t$ is the target year of yield prediction, then in the test phase, the average yield in year $t$, $\hat{\overline{Y}}_t$, can be substituted with $\overline{Y}_{t-1}$, and the unobserved portion of weather data in $W_t$ can be substituted with the predicted weather data. In the training phase, however, such substitution is unnecessary since all the input data is available.

# DESIGN OF EXPERIMENTS

We used the following hyperparameters to train the CNN-RNN model. The W-CNN and S-CNN models both have four convolutional layers with detailed structure provided in **Table S2** in the **Supplementary Material**. In the CNN models, downsampling was performed by average pooling with stride of 2. The output of W-CNN is followed by a fully connected layer, which has 60 neurons for corn yield prediction and 40 neurons for soybean yield prediction. The output of the S-CNN model is followed by a fully connected layer that has 40 neurons. The RNN layer has a time length of 5 years since we considered a 5-year yield dependencies. The RNN layer has LSTM cells with 64 hidden units. After trying different network designs, we found this architecture to provide the best overall performance.

All weights were initialized with the Xavier method (Glorot and Bengio, 2010). We used stochastic gradient decent (SGD) with mini-batch size of 25. Adam optimizer (Kingma and Ba, 2014) was used with the learning rate of 0.03%, which was divided by 2 every 60,000 iterations. The model was trained for the maximum 350,000 iterations. We used rectified linear unit (ReLU) activation function for the CNNs and FC layer. The output layer had a linear activation function. The proposed model was implemented in Python using the Tensorflow library (Abadi et al., 2016), and training time took about an hour on a CPU (i7-4790, 3.6 GHz).

**Figure S2** in the **Supplementary Material** shows the plots of training loss and validation loss for corn and soybean yield prediction.

We also implemented three other popular prediction models for comparison: random forest (RF) (Breiman, 2001), deep fully connected neural network (DFNN) (Khaki and Wang, 2019), and the least absolute shrinkage and selection operator (LASSO) (Tibshirani, 1996). Implementation details of these models are provided as follows.

- The random forest is a powerful non-parametric model that uses ensemble learning to avoid overfitting. We tried different numbers of trees and found that 50 trees resulted in the most accurate predictions. Increasing the number of trees in the RF model increased the training time without improving the accuracy of results. We also tried different numbers of maximum depth of the trees. We found that the maximum depth of the tree equal to 10 led to the most accurate predictions. Increasing the maximum depth of the trees resulted in overfitting while decreasing the maximum depth of the trees resulted in the low prediction accuracy.
- The DFNN model with multiple stacked nonlinear layers is a powerful nonlinear model. The DFNN model had 9 layers and 50 neurons in each layer as in Khaki and Wang (2019). The DFNN model used the state-of-the-art deep learning techniques such as batch normalization (Ioffe and Szegedy, 2015) and residual learning (He et al., 2016) to increase the prediction accuracy. We tried different numbers of hidden layers and found that nine hidden layers led to the most accurate predictions.
- LASSO is used as a benchmark model to compare the linear and nonlinear effects of soil and weather data in the yield prediction. Different values for coefficient of $L_1$ term (Ng, 2004) in LASSO model were tried and we found that values between 0.3 and 0.5 led to the most accurate predictions.

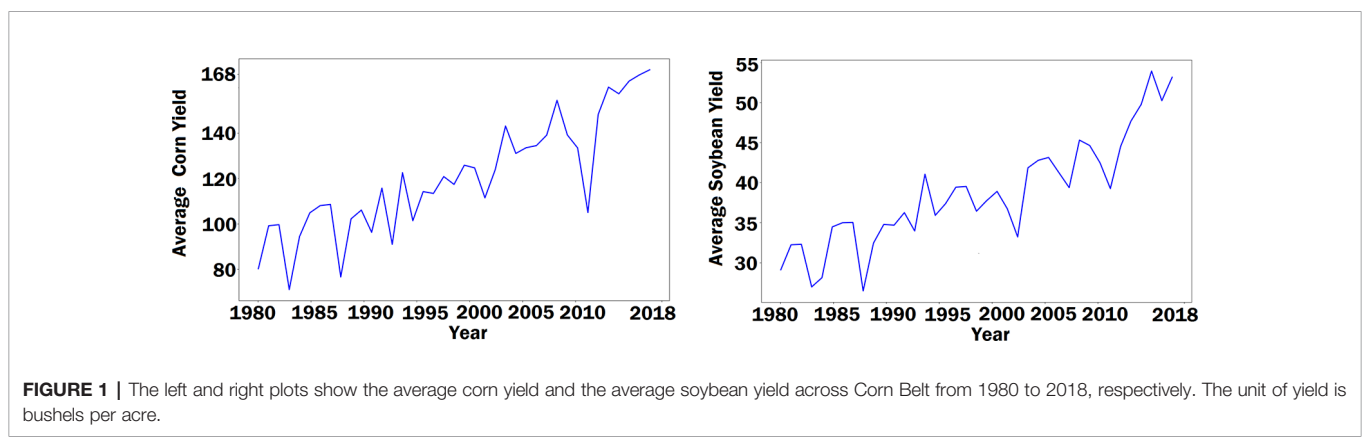

**FIGURE 1** | The left and right plots show the average corn yield and the average soybean yield across Corn Belt from 1980 to 2018, respectively. The unit of yield is bushels per acre.

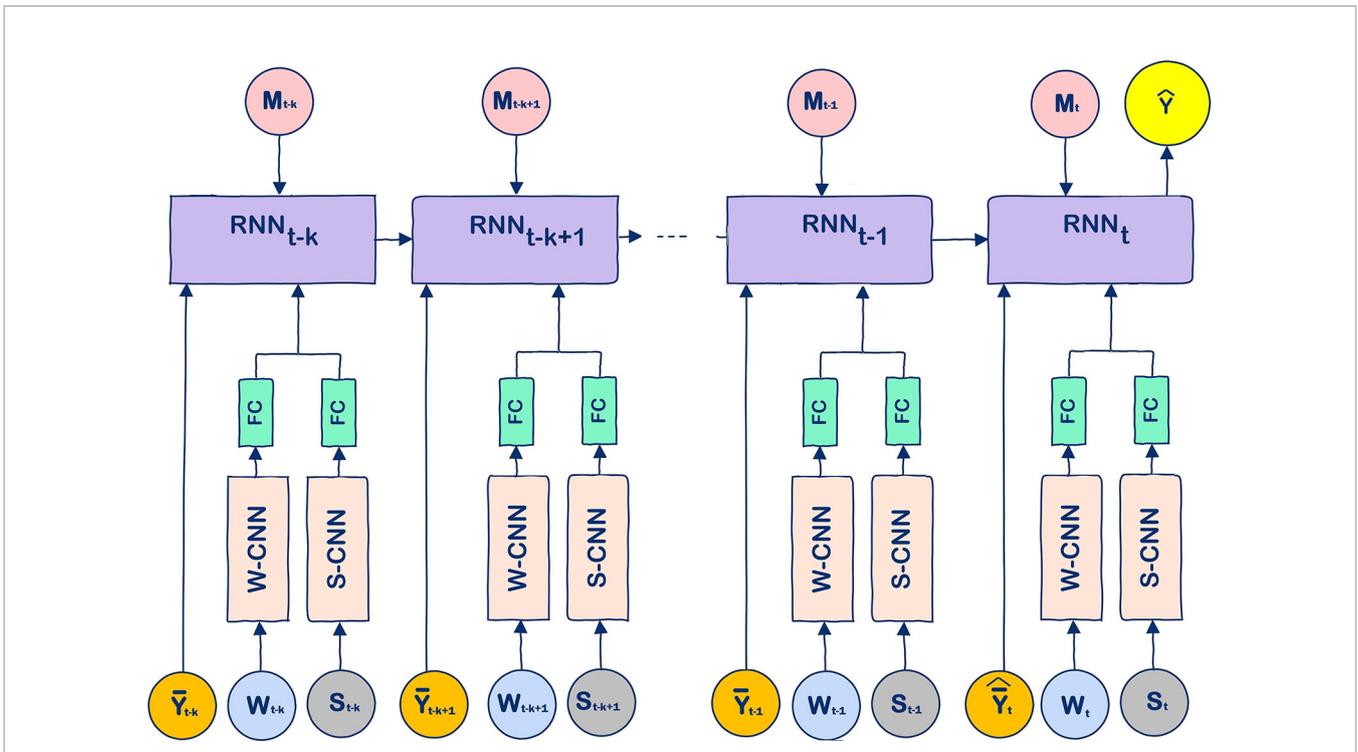

**FIGURE 2** | The unrolled modeling structure of the proposed CNN-RNN model. Input variables $W_t$, $S_t$, $\overline{Y}_t$, $\hat{\overline{Y}}_t$, and $M_t$ denote the weather, soil, average yield, predicted average yield, and management data at time step $t$, respectively, and $k$ denotes the length of time dependencies. At the test phase, $\overline{Y}_{t-1}$ was used as an estimator for $\hat{\overline{Y}}_t$ in this paper.

## RESULTS

We implemented all four models in the most efficient manner to predict corn and soybean yields. We took three years, namely 2016, 2017, and 2018, as validation years and predicted the yields of corn and soybean for these years. For each validation year, training data included data from 1980 to the year before the corresponding validation year. **Tables 1** and **2** compare the performances of the four models on both training and validation datasets with respect to the RMSE and correlation coefficient for corn and soybean yield prediction, respectively. **Table 3** shows the summary statistics for the validation years.

**TABLE 1** | Corn yield prediction performance for years 2016, 2017, and 2018.

| Model | Validation Year | Training RMSE | Training Correlation Coefficient (%) | Validation RMSE | Validation Correlation Coefficient (%) |
|---|---|---|---|---|---|
| CNN-RNN | 2016 | 13.26 | 93.02 | 16.48 | 85.82 |
|  | 2017 | 12.75 | 93.68 | 15.74 | 88.24 |
|  | 2018 | 11.48 | 94.99 | 17.64 | 87.82 |
| RF | 2016 | 13.38 | 92.74 | 25.48 | 69.52 |
|  | 2017 | 14.31 | 92.39 | 29.40 | 69.03 |
|  | 2018 | 14.40 | 92.39 | 26.02 | 70.55 |
| DFNN | 2016 | 12.34 | 94.43 | 27.23 | 81.91 |
|  | 2017 | 11.21 | 95.09 | 23.88 | 79.57 |
|  | 2018 | 11.54 | 95.25 | 21.37 | 79.85 |
| LASSO | 2016 | 19.88 | 81.81 | 32.58 | 61.90 |
|  | 2017 | 20.62 | 81.83 | 27.06 | 61.18 |
|  | 2018 | 20.81 | 83.63 | 31.30 | 55.95 |

*RF and DFNN stand for random forest and deep fully connected neural network, respectively. The average ± standard deviation for corn yield in years 2016, 2017, and 2018 are, respectively, 165.72 ± 30.35, 168.50 ± 32.88, and 170.77± 34.95. The unit of RMSE is bushels per acre.*

TABLE 2 | Soybean yield prediction performance for years 2016, 2017, and 2018.

| Model | Validation Year | Training RMSE | Training Correlation Coefficient (%) | Validation RMSE | Validation Correlation Coefficient (%) |
|---|---|---|---|---|---|
| CNN-RNN | 2016 | 3.38 | 94.36 | 4.15 | 85.45 |
| | 2017 | 3.08 | 95.35 | 4.32 | 87.08 |
| | 2018 | 3.85 | 92.54 | 4.91 | 87.09 |
| RF | 2016 | 4.29 | 90.77 | 8.69 | 68.60 |
| | 2017 | 4.38 | 90.75 | 8.61 | 32.90 |
| | 2018 | 4.39 | 90.92 | 12.78 | 40.82 |
| DFNN | 2016 | 4.06 | 91.29 | 7.51 | 72.83 |
| | 2017 | 4.35 | 90.36 | 6.25 | 72.07 |
| | 2018 | 3.98 | 92.09 | 5.89 | 79.78 |
| LASSO | 2016 | 6.49 | 75.86 | 8.05 | 51.03 |
| | 2017 | 6.49 | 77.22 | 7.66 | 60.30 |
| | 2018 | 6.51 | 77.78 | 9.49 | 64.74 |

*RF and DFNN stand for random forest and deep fully connected neural network, respectively. The average ± standard deviation for corn yield in years 2016, 2017, and 2018 are, respectively, 53.94 ± 7.23, 50.24 ± 8.72, and 53.17± 9.72. The unit of RMSE is bushels per acre.*

TABLE 3 | Summary statistics of validation data. The unit of yield is bushels per acre.

| Response | Validation Year | Mean | Standard Deviation | Number of Locations with Available Ground Truth Yield |
|---|---|---|---|---|
| Corn | 2016 | 165.72 | 30.35 | 914 |
| | 2017 | 168.50 | 32.88 | 910 |
| | 2018 | 170.77 | 34.95 | 807 |
| Soybean | 2016 | 53.94 | 7.23 | 798 |
| | 2017 | 50.24 | 8.72 | 800 |
| | 2018 | 53.17 | 9.72 | 684 |

The results suggest that the hybrid CNN-RNN model significantly outperformed the other three models to varying extent. The weak performance of LASSO was mainly due to its linear property, which could not capture the nonlinear effects of soil conditions and weather components. The DFNN performed better than LASSO since DFNN was able to capture the nonlinear effects of environmental components. DFNN had a better performance compared to the RF model with respect to all performance measures except the validation RMSE of the year 2016 for corn yield prediction. RF showed a better performance compared to the LASSO with respect to all performance measures except the validation RMSE of the year 2017 for corn yield prediction. LASSO achieved better performance compared to the RF with respect to all performance measures except correlation coefficient of the year 2016 for soybean yield prediction. The CNN-RNN model outperformed all other three models with respect to all measures for all three validation years. The CNN-RNN was effective in predicting yields of both corn and soybean with RMSE for the validation data being approximately 9% and 8% of their respective average values.

The reasons for the outstanding performance of the CNN-RNN model are as follows: (1) the RNN part of the CNN-RNN model considered the genetic improvements of the seeds by capturing the yearly time dependencies of the yield, (2) the W-CNN part of the CNN-RNN model captured the internal time dependencies of the weather data, (3) the S-CNN part of the CNN-RNN model considered the spatial dependencies of soil data measured at different depths underground, and (4) the CNN-RNN model took into account the nonlinear effects of environmental components.

The results also suggest that the CNN-RNN model had a consistent performance for both corn and soybean yield predictions across all validation years. To examine the yield prediction error for individual counties, we obtained the absolute prediction errors of the year 2018 for corn and soybean yield predictions. **Figure 3** shows the prediction error maps for corn and soybean, respectively. As shown in **Figure 3**, prediction errors were consistently low for most of the counties.

**Figure S3** in the **Supplementary Material** shows the plots of predicted yield versus the ground truth yield for the 2018 validation year for corn and soybean yield predictions, respectively. To see whether the CNN-RNN model can preserve some of the distributional properties of the ground truth yield, we plotted the probability density functions of the ground truth yield and the predicted yield by the CNN-RNN model. As shown in **Figure S4** in the **Supplementary Material**, the CNN-RNN model can approximately preserve some of the distributional properties of the ground truth yield.

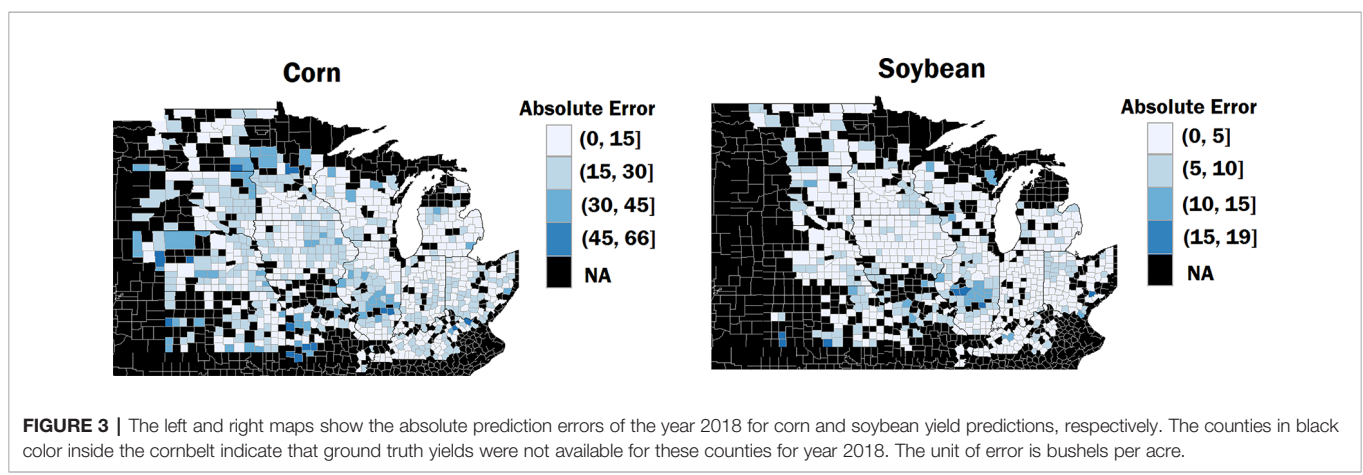

FIGURE 3 | The left and right maps show the absolute prediction errors of the year 2018 for corn and soybean yield predictions, respectively. The counties in black color inside the cornbelt indicate that ground truth yields were not available for these counties for year 2018. The unit of error is bushels per acre.

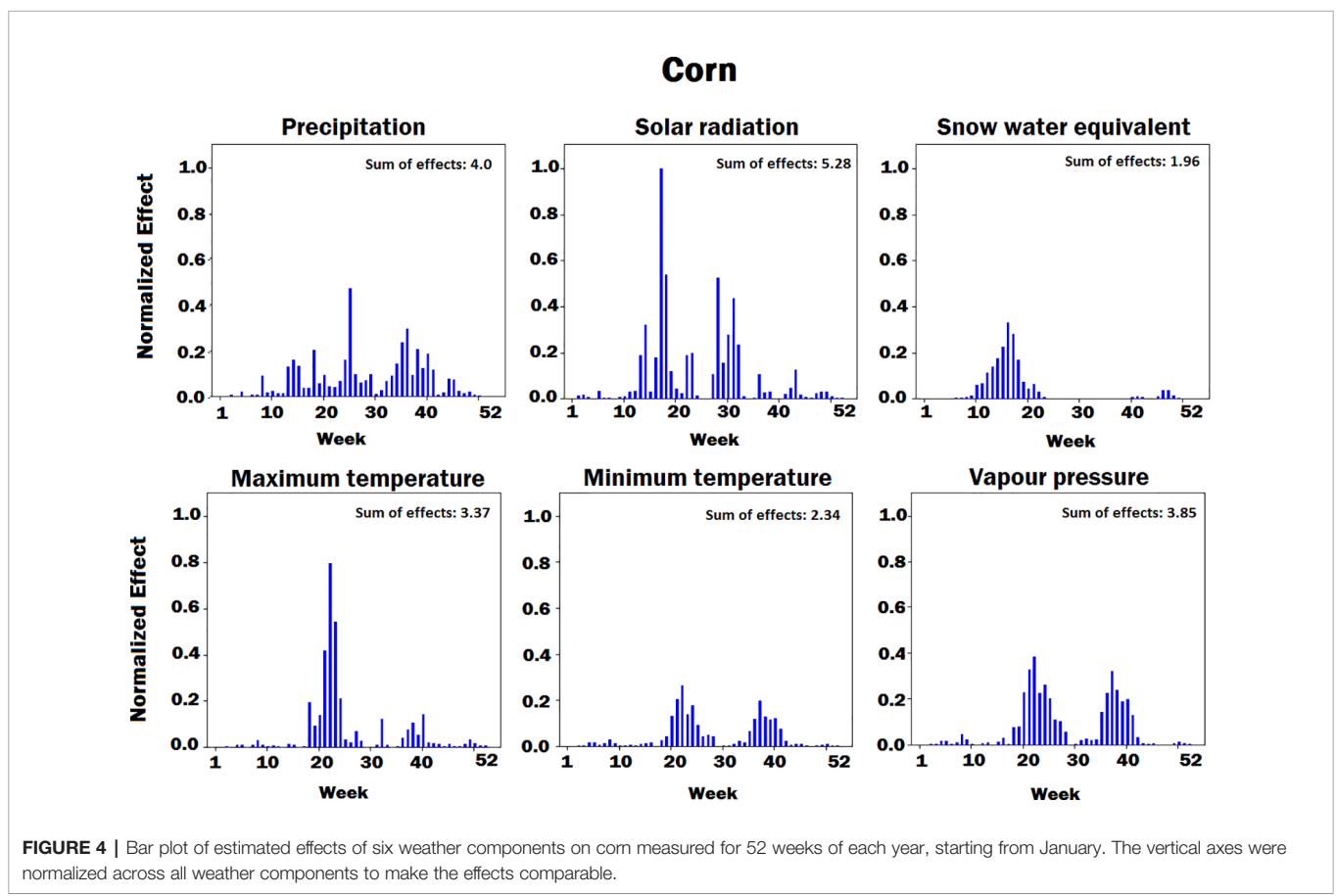

**FIGURE 4** | Bar plot of estimated effects of six weather components on corn measured for 52 weeks of each year, starting from January. The vertical axes were normalized across all weather components to make the effects comparable.

## ANALYSIS

### Feature Selection

We have predicted the crop yield based on the input variables such as weather components, soil conditions, and management practices. To find the relative importance of each factor, we performed feature selection based on the trained CNN-RNN model. We performed two feature selections, one for corn and one for soybean yield prediction. We used guided backpropagation method to backpropagate the positive gradients to find input variables, which maximize the activation of our targeted neurons (Springenberg et al., 2014; Khaki and Wang, 2019). First, we fed all validation samples to the CNN-RNN model and computed the average activation of all neurons in the output of the RNN cell at time step $t$. We set the gradient of activated neurons to be 1 and the other neurons to be 0. Then, we backpropagated the gradients of the activated neurons to the input space to find the important input variables based on the magnitude of the gradient (the bigger, the more important).

**Figures 4–8** illustrate the estimated effects of weather components, soil conditions measured at the soil surface and also at different depths, and management practices. The effects were normalized within each group, namely, weather components, soil conditions, and management practices, to make the effects comparable.

The innovative aspect of our importance analysis is the temporal resolution that allows identification of critical periods towards a deeper understanding of how the complex agronomic system works (**Figures 4** and **5**). Among six weather variables examined here, solar radiation was the most sensitive factor and snow the least sensitive factor in the corn yield factor. This is reasonable from an agronomic perspective as radiation is the key driver of photosynthesis and subsequently biomass production and grain yield (Sinclair and Horie, 1989). On the other hand, snow-water has an effect on soil water balance as does precipitation and vapor deficit (Ritchie, 1998), which were more important variables than snow because their effect on crop yield lasts longer and especially during summer time. Snow is somewhat important before and after the growing season, which was well depicted by the current analysis (**Figures 4** and **5**). Radiation showed two picks (very sensitive periods for yield prediction), one around week 15, which is prior to crop planting, and one around week 30, which coincides with the most critical corn stage (silking). Around silking time, the kernel number per plant is determined and literature shows that there is a very strong relationship between the kernel number (main determinant of grain yield) and the plant growth rate that is basically driven by photosynthesis (Andrade et al., 1999); thus, the current analysis captured that phenomenon. Maximum temperature was found most sensitive in yield prediction around week 20, a period that typically coincides with corn planting (May 13). From an agronomic perspective, temperature is very important at that period because it affects seed germination and emergence, and typically high temperature results in fast and uniform emergence while low temperatures in slow and non-uniform emergence that affect plant growth yield. The minimum temperature was most sensitive around weeks 22 (May 30) and 35 (August 26). The importance of week 22 is the same as with maximum temperature for seed emergence. Interestingly, the model picked the importance of minimum temperature during grain fill period, which is well known to affect corn yields (Hatfield and Prueger, 2015; Schauberger et al., 2017). The same is also true for soybeans, and this was captured by the model. Another interesting result from this analysis is the increasing importance of precipitation during grain fill period (weeks 30 to 40), which also agrees with experimental studies (Hatfield et al., 2011; Hatfield et al., 2018).

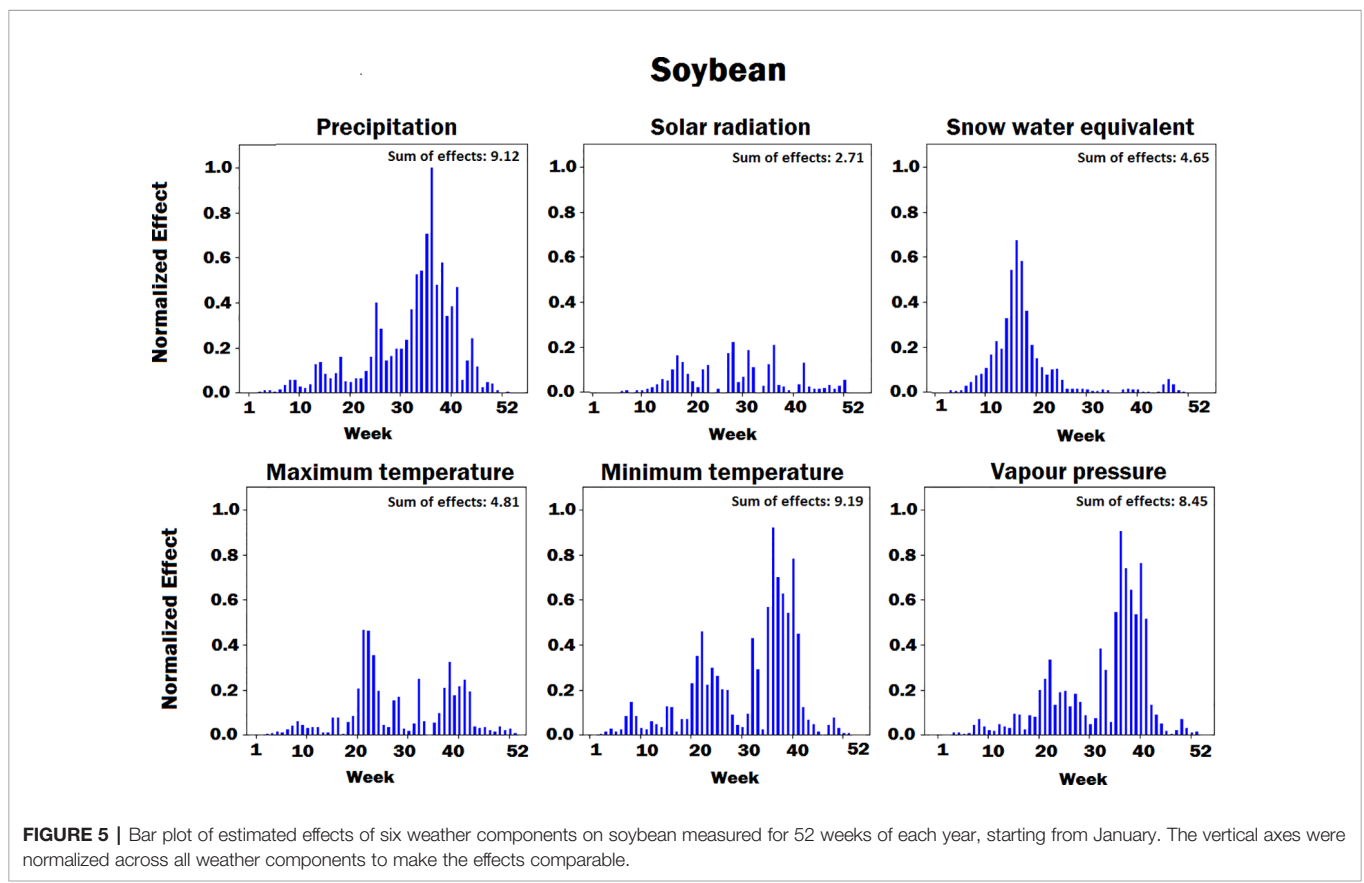

**FIGURE 5 |** Bar plot of estimated effects of six weather components on soybean measured for 52 weeks of each year, starting from January. The vertical axes were normalized across all weather components to make the effects comparable.

In terms of soil variables (see **Figures 6** and **7**), our analysis showed several factors to be sensitive to both yield predictions. Explaining all these factors is agronomically beyond the scope of this paper, but all of these factors are known to affect soil water and nitrogen supply to the crop and thus crop yields (Archontoulis et al., 2016).

In terms of planting dates, corn yield prediction was least sensitive around April 20th to May 15th, a period that is regarded as the optimum planting date for corn in the Corn Belt region (Baum et al., 2018). From an agronomic perspective, corn yield decreases outside this optimum range and the model was able to capture this fact by increasing sensitivity to planting date. Opposite results were obtained for soybean, whereas the models were most sensitive during May 15 to end of May period, which is regarded as the optimum planting time for soybean (Egli and Bruening, 1992).

To evaluate the performance of the feature selection method, we obtained prediction results based on a subset of features. We trained the CNN-RNN model on the data from 1980 to 2016 and used 2017 data to do feature selection. Finally, we evaluated the performance of feature selection method on the 2018 yield prediction. We sorted all the features based on their estimated effects and selected the 50% and 75% most important features. **Table 4** shows the yield prediction performance of the CNN-RNN model using these selected features. The prediction accuracy of the CNN-RNN model did not drop significantly compared to the CNN-RNN model using all the features (100%), which suggested that the feature selection method can successfully find the important features.

## Importance Comparison Between Environment and Management Practices

To compare the individual importance of weather components, soil conditions, and management practices, we performed the yield prediction using the following models:

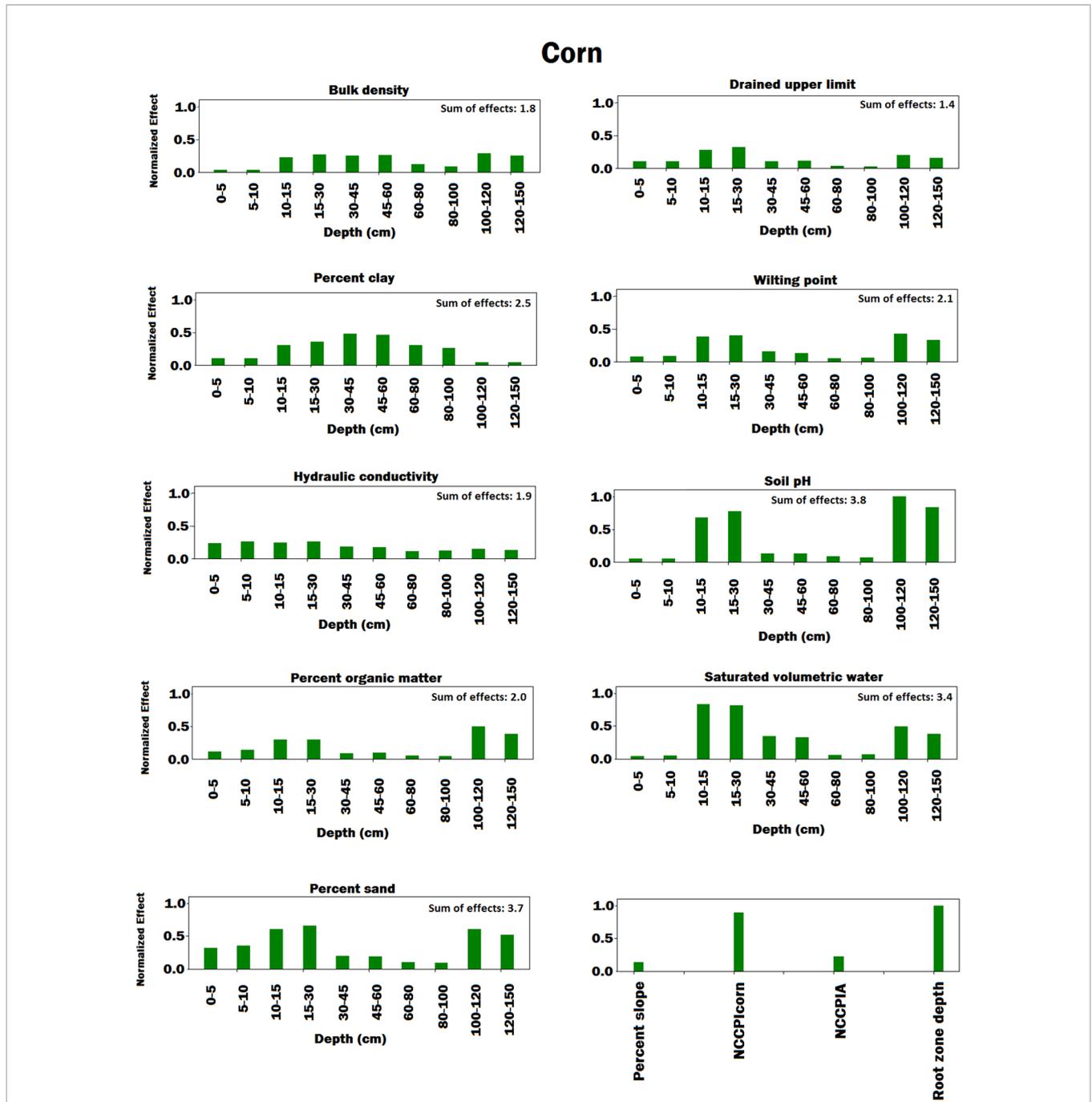

**FIGURE 6** | Bar plot of estimated effects of soil conditions measured at different depths of soil and soil conditions measured at the soil surface on corn. The vertical axes were normalized across soil conditions to make the effects comparable. Separate normalizations were done for soil conditions measured at different depths of soil and soil conditions measured at the soil surface. The NCCPIcorn and NCCPIA stand for national commodity crop productivity index for corn and national commodity crop productivity index for all crops, respectively.

**CNN-RNN(W)**: this model uses the CNN-RNN model to predict yield based on the weather data without using soil and management data. This model only captures the linear and nonlinear effects of weather data.

**CNN-RNN(S)**: this model uses the CNN-RNN model to predict yield based on the soil data without using weather and management data. This model only captures the linear and nonlinear effects of soil data.

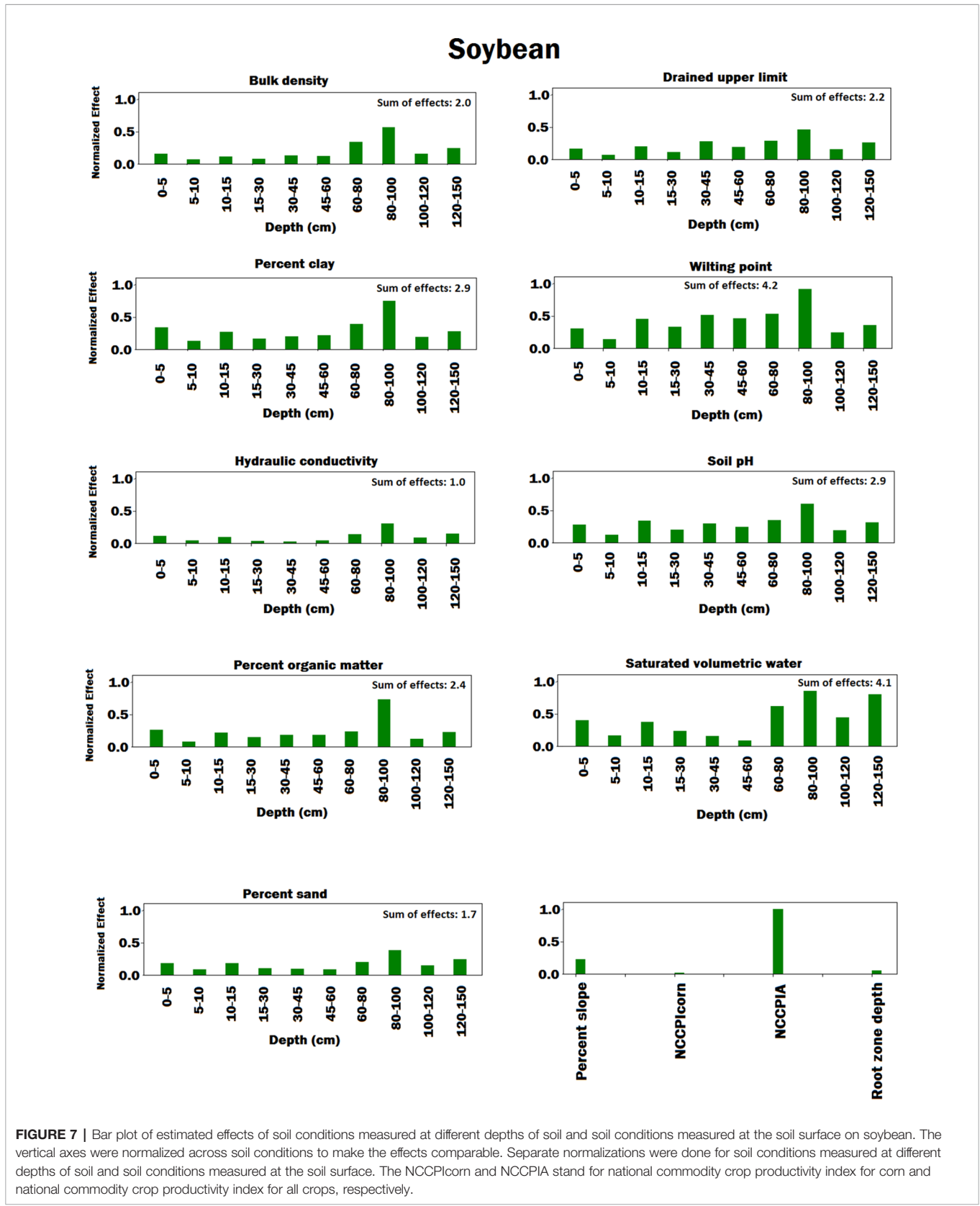

FIGURE 7 | Bar plot of estimated effects of soil conditions measured at different depths of soil and soil conditions measured at the soil surface on soybean. The vertical axes were normalized across soil conditions to make the effects comparable. Separate normalizations were done for soil conditions measured at different depths of soil and soil conditions measured at the soil surface. The NCCPIcorn and NCCPIA stand for national commodity crop productivity index for corn and national commodity crop productivity index for all crops, respectively.

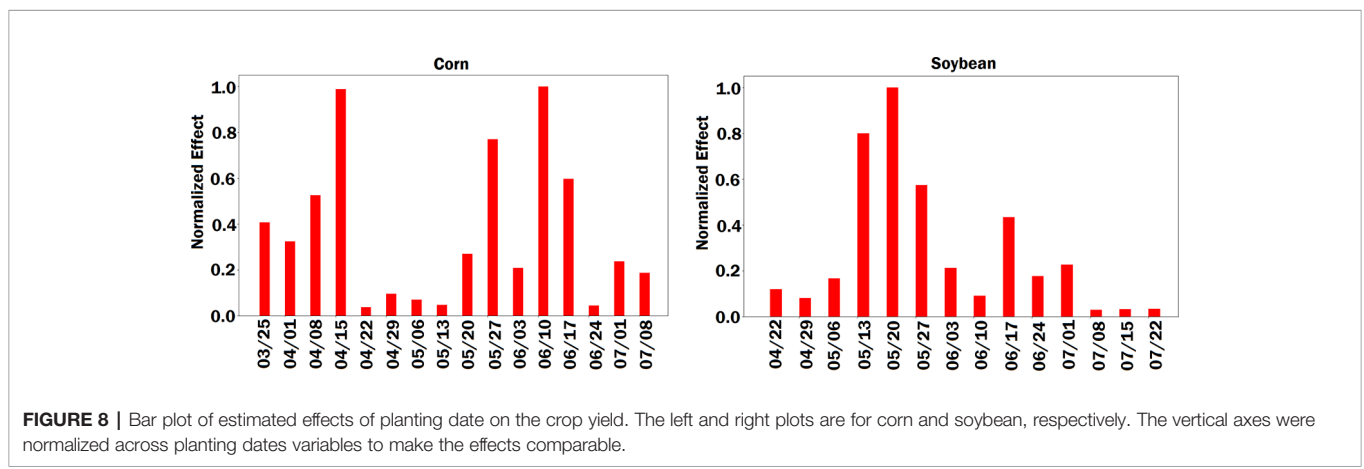

FIGURE 8 | Bar plot of estimated effects of planting date on the crop yield. The left and right plots are for corn and soybean, respectively. The vertical axes were normalized across planting dates variables to make the effects comparable.

- **CNN-RNN(M)**: this model uses the CNN-RNN model to predict yield based on the management data without using weather and soil data. This model only captures the linear and nonlinear effects of management data.
- **Average**: this model provides a benchmark using only the average of yield for prediction.

**Table 5** compares the performance of the above four models in the yield prediction of corn and soybean CNN-RNN(W) demonstrated comparable performance with CNN-RNN(S) for both corn and soybean yield prediction, and their prediction accuracies were significantly higher than CNN-RNN(M). The results suggested that weather and soil were equally important factors in the yield prediction, and explained more of the variation in the crop yield than management practices (planting dates). The results also revealed that planting dates had more effects on the soybean compared to the corn.

## Generalization Power of the CNN-RNN Model

To examine the power of the model in generalizing the prediction to the locations that have never been tested, we randomly excluded locations from the training data (1980–2017) and trained the CNN-RNN model on the remaining locations. Then, we tested the model on the excluded locations for the 2018 yield prediction. We used k-fold cross-validation for estimating the generalization power of the proposed model. The number of locations with available 2018 ground truth yield are 807 and 684 for corn and soybean, respectively. We applied 5-fold cross-validation on locations with available 2018 ground truth yield, which resulted in having 163 and 140 locations in each fold for corn and soybean, respectively.

**Table 6** shows the 5-fold cross-validation performance of the CNN-RNN model for corn and soybean yield predictions. As shown in **Table 6**, the prediction accuracy of the CNN-RNN model did not deteriorate considerably compared to the corresponding results in **Tables 1** and **2**, which demonstrate that the CNN-RNN model can successfully generalize the yield prediction to untested locations. **Figure S5** in the **Supplementary Material** shows the performance of the CNN-RNN model on the individual 100 untested locations. As shown in **Figure S5** in the **Supplementary Material**, the prediction errors were consistently low for most of the untested locations.

TABLE 4 | Yield prediction performance of the CNN-RNN model on the subset of features. The unit of RMSE is bushels per acre.

| Response | Model | Training RMSE | Training Correlation Coefficient (%) | Validation RMSE | Validation Correlation Coefficient (%) |
|---|---|---|---|---|---|
| Corn | CNN-RNN using 50% of features | 16.86 | 89.69 | 22.24 | 80.31 |
|  | CNN-RNN using 75% of features | 15.0 | 91.76 | 21.65 | 79.76 |
|  | CNN-RNN using 100% of features | 11.48 | 94.99 | 17.64 | 87.82 |
| Soybean | CNN-RNN using 50% of features | 4.97 | 87.44 | 5.95 | 79.77 |
|  | CNN-RNN using 75% of features | 4.04 | 92.15 | 5.36 | 83.67 |
|  | CNN-RNN using 100% of features | 3.85 | 92.54 | 4.91 | 87.09 |

TABLE 5 | Yield prediction performances of CNN-RNN(W), CNN-RNN(S), CNN-RNN(M), and Average model. Validation year is 2018. The unit of RMSE is bushels per acre.

| Response | Model | Training RMSE | Training Correlation Coefficient (%) | Validation RMSE | Validation Correlation Coefficient (%) |
|---|---|---|---|---|---|
| Corn | CNN-RNN(W) | 17.12 | 89.00 | 24.75 | 72.15 |
|  | CNN-RNN(S) | 19.13 | 84.70 | 24.41 | 72.82 |
|  | CNN-RNN(M) | 26.77 | 69.49 | 33.03 | 32.99 |
|  | Average | 37.51 | 0.0 | 35.56 | 0.0 |
| Soybean | CNN-RNN(W) | 4.59 | 89.62 | 5.95 | 78.79 |
|  | CNN-RNN(S) | 5.60 | 83.25 | 5.93 | 81.23 |
|  | CNN-RNN(M) | 7.86 | 59.86 | 8.67 | 48.78 |
|  | Average | 10.28 | 0.0 | 10.10 | 0.0 |

TABLE 6 | 5-fold cross-validation performance of generalization power of the CNN-RNN model for corn and soybean prediction. The unit of RMSE is bushels per acre.

| Response Variable | Training RMSE | Training Correlation (%) | Validation RMSE | Validation Correlation (%) |
|---|---|---|---|---|
| Corn yield | 14.49 | 92.0 | 24.10 | 75.04 |
| Soybean yield | 4.21 | 91.19 | 6.35 | 77.84 |

## Yield Prediction Using Predicted Weather Data

Weather is one of the important factors in the crop yield prediction but it is unknown *a priori*. As such, weather prediction is an unavoidable part of crop yield prediction. To evaluate the impacts of weather prediction on the performance of the CNN-RNN model, we obtained the 2018 yield prediction results for state of Iowa using predicted weather data. We took 2017 weather data from June to September as the predicted weather data for the corresponding time interval of the year 2018. We compared the RMSE and the predicted state average yield of the CNN-RNN model using perfect weather data with the RMSE and the predicted state average yield of the CNN-RNN model using predicted weather data. To better see the effect of weather prediction on the yield prediction, we updated the predicted weather data every week with its corresponding ground truth 2018 weather data starting from June until September, and prediction results were obtained for each week. As shown in **Figure 9**, the more we updated the predicted weather data with the ground truth weather data every week, the more prediction error decreased, which revealed how sensitive yield prediction is to weather prediction. The results also suggested that a perfect weather prediction model could considerably improve the yield prediction results.

## CONCLUSION

In this paper, we presented a deep learning-based approach for crop yield prediction, which accurately predicted corn and soybean yields across the entire Corn Belt in the United States based on environmental data and management practices. Most importantly, our methodology moved beyond prediction, as it provided key results towards explaining yield prediction (variable importance by time period).

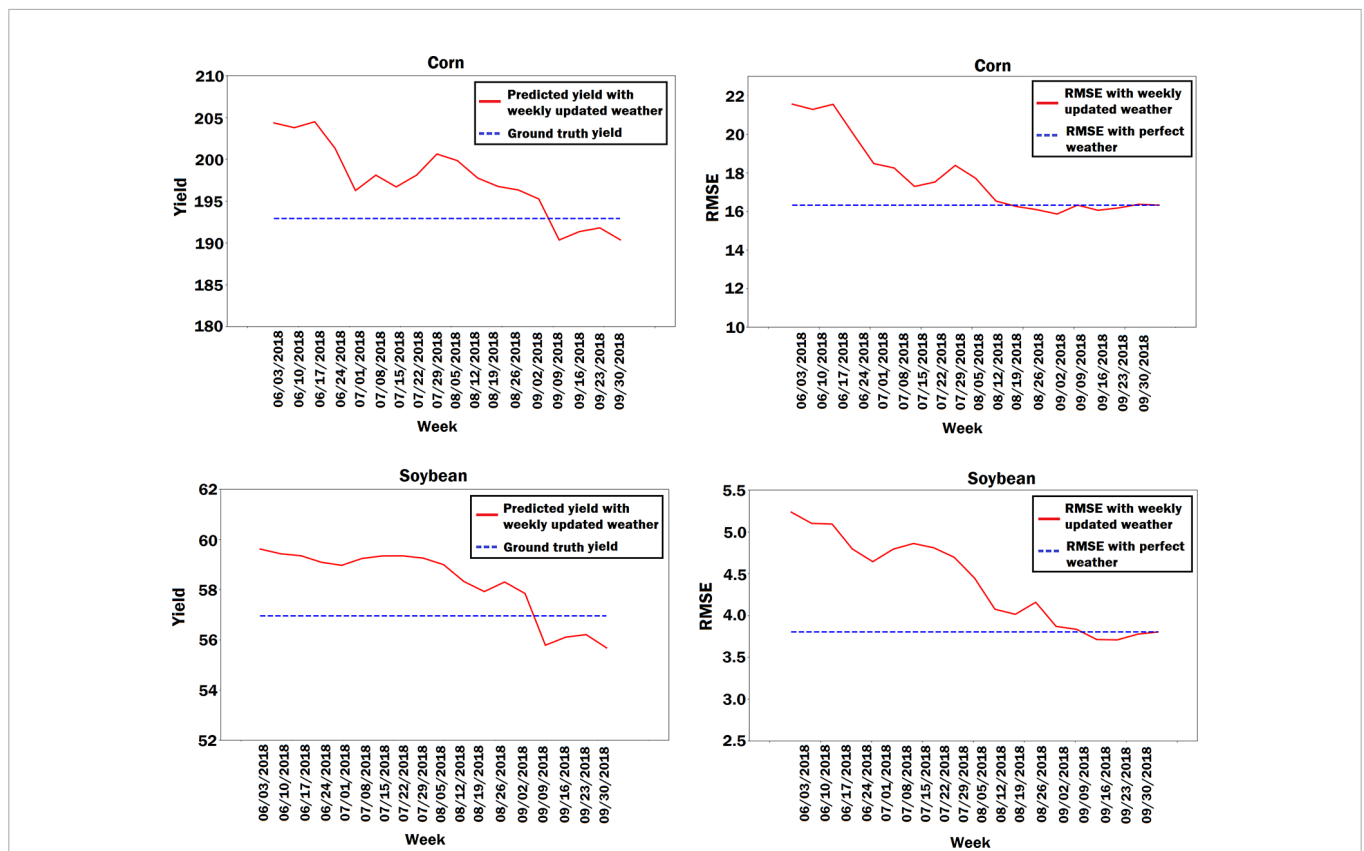

FIGURE 9 | Plots of predicted state average yield and RMSE based on the predicted weather data for the state of Iowa. The predicted weather data was updated every week with its corresponding ground truth 2018 weather data and prediction results were obtained for each week. The units of yield and RMSE are bushels per acre.

The proposed method significantly outperformed other popular methods such as LASSO, random forest, and DFNN. The proposed model is a hybrid one that combines CNNs and RNNs. The CNN part of the model was designed to capture the internal temporal dependencies of weather data and the spatial dependencies of soil data measured at different depths underground. The RNN part of the model was designed to capture the increasing trend of crop yield over years due to continuous improvement in plant breeding and management practices. The performance of the model was relatively sensitive to many variables, including weather, soil, and management. The proposed model successfully predicted yields in untested environments; thus, it could be used in future yield prediction tasks.

One of the principal limitations of deep learning models is their black box property. To make the proposed model less of a black box and more explainable, feature selection was performed based on the trained CNN-RNN model using the backpropagation method. The feature selection method successfully estimated the individual effect of weather components, soil conditions, and management variables as well as the time period that these variables become important, which is an innovation of this study. This method could be extended to address other research problems. For example, similar approaches could be used to classify a hybrid as either low-yielding or high-yielding based on its relative performance against other hybrids at the same location.

## DATA AVAILABILITY STATEMENT

The data analyzed in this study are publicly available and are referenced in the section *Data* of the paper. The source code of the CNN-RNN model is available on GitHub (Khaki, 2019).

## AUTHOR CONTRIBUTIONS

SK, LW, and SA conceived the study. SK implemented the computational experiments. SK, LW, and SA wrote the paper.

## FUNDING

This work was partially supported by the National Science Foundation under the LEAP HI and GOALI programs (grant number 1830478) and under the EAGER program (grant number 1842097).

## SUPPLEMENTARY MATERIALS

The Supplementary Material for this article can be found online at: https://www.frontiersin.org/articles/10.3389/fpls.2019.01750/full#supplementary-material

## REFERENCES

Abadi, M., Barham, P., Chen, J., Chen, Z., Davis, A., Dean, J., et al. (2016). "TensorFlow: A system for large scale machine learning," in *12th USENIX Symposium on Operating Systems Design and Implementation* (Savannah, GA, USA), vol. 16. 265–283.

Andrade, F. H., Vega, C., Uhart, S., Cirilo, A., Cantarero, M., and Valentinuz, O. (1999). Kernel number determination in maize. *Crop Sci.* 39, 453–459. doi: 10.2135/cropsci1999.0011183x0039000200026x

Archontoulis, S. V., Huber, I., Miguez, F. E., Thorburn, P. J., Rogovska, N., and Laird, D. A. (2016). A model for mechanistic and system assessments of biochar effects on soils and crops and trade-offs. *GCB Bioenergy* 8, 1028–1045. doi: 10.1111/gcbb.12314

Awad, M. M. (2019). Toward precision in crop yield estimation using remote sensing and optimization techniques. *Agriculture* 9, 54. doi: 10.3390/agriculture9030054

Baum, M., Archontoulis, S., and Licht, M. (2018). Planting date, hybrid maturity, and weather effects on maize yield and crop stage. *Agron. J.* 111, 1–11. doi: 10.2134/agronj2018.04.0297

Bengio, Y., Simard, P., and Frasconi, P. (1994). Learning long-term dependencies with gradient descent is difficult. *IEEE Trans. Neural Networks* 5, 157–166. doi: 10.1109/72.279181

Borovykh, A., Bohte, S., and Oosterlee, C. W. (2017). Conditional time series forecasting with convolutional neural networks. *arXiv preprint arXiv:1703.04691*.

Breiman, L. (2001). Random forests. *Mach. Learn.* 45, 5–32. doi: 10.1023/A:1010933404324

Drummond, S. T., Sudduth, K. A., Joshi, A., Birrell, S. J., and Kitchen, N. R. (2003). Statistical and neural methods for site–specific yield prediction. *Trans. ASAE* 46, 5. doi: 10.13031/2013.12541

Egli, D., and Bruening, W. (1992). Planting date and soybean yield: evaluation of environmental effects with a crop simulation model: Soygro. *Agric. For. Meteorol.* 62, 19–29. doi: 10.1016/0168-1923(92)90003-m

Fukuda, S., Spreer, W., Yasunaga, E., Yuge, K., Sardsud, V., and Müller, J. (2013). Random forests modelling for the estimation of mango (mangifera indica l. cv. chok anan) fruit yields under different irrigation regimes. *Agric. Water Manage.* 116, 142–150. doi: 10.1016/j.agwat.2012.07.003

Glorot, X., and Bengio, Y. (2010). "Understanding the difficulty of training deep feedforward neural networks," in *Proceedings of the Thirteenth International Conference on Artificial Intelligence and Statistics* (Sardinia, Italy). 249–256.

Goodfellow, I., Bengio, Y., and Courville, A. (2016). *Deep Learning* Vol. 1 (Cambridge: MIT Press).

gSSURGO (2019). *Soil Survey Staff. Gridded Soil Survey Geographic (gSSURGO) Database for the United States of America and the Territories, Commonwealths, and Island Nations served by the USDA-NRCS* (United States Department of Agriculture, Natural Resources Conservation Service).

Hatfield, J. L., and Prueger, J. H. (2015). Temperature extremes: Effect on plant growth and development. *Weather Climate Extremes* 10, 4–10. doi: 10.1016/j.wace.2015.08.001

Hatfield, J. L., Boote, K. J., Kimball, B. A., Ziska, L., Izaurralde, R. C., Ort, D., et al. (2011). Climate impacts on agriculture: implications for crop production. *Agron. J.* 103, 351–370. doi: 10.2134/agronj2010.0303

Hatfield, J., Wright-Morton, L., and Hall, B. (2018). Vulnerability of grain crops and croplands in the midwest to climatic variability and adaptation strategies. *Clim. Change* 146, 263–275. doi: 10.1007/s10584-017-1997-x

He, K., Zhang, X., Ren, S., and Sun, J. (2016). "Deep residual learning for image recognition," in *Proceedings of the IEEE Conference on Computer Vision and Pattern Recognition*, Las Vegas, USA. 770–778.

Hochreiter, S., and Schmidhuber, J. (1997). Long short-term memory. *Neural Comput.* 9, 1735–1780. doi: 10.1162/neco.1997.9.8.1735

Horie, T., Yajima, M., and Nakagawa, H. (1992). Yield forecasting. *Agric. Syst.* 40, 211–236. doi: 10.1016/0308-521X(92)90022-G

Hornik, K., Stinchcombe, M., and White, H. (1989). Multilayer feedforward networks are universal approximators. *Neural Networks* 2, 359–366. doi: 10.1016/0893-6080(89)90020-8

Ince, T., Kiranyaz, S., Eren, L., Askar, M., and Gabbouj, M. (2016). Real-time motor fault detection by 1-d convolutional neural networks. *IEEE Trans. Ind. Electron.* 63, 7067–7075. doi: 10.1109/tie.2016.2582729

Ioffe, S., and Szegedy, C. (2015). Batch normalization: Accelerating deep network training by reducing internal covariate shift. *arXiv preprint arXiv:1502.03167*.

Jeong, J. H., Resop, J. P., Mueller, N. D., Fleisher, D. H., Yun, K., Butler, E. E., et al. (2016). Random forests for global and regional crop yield predictions. *PloS One* 11, e0156571. doi: 10.1371/journal.pone.0156571

Jiang, D., Yang, X., Clinton, N., and Wang, N. (2004). An artificial neural network model for estimating crop yields using remotely sensed information. *Int. J. Remote Sens.* 25, 1723–1732. doi: 10.1080/0143116031000150068

Khaki, S., and Khalilzadeh, Z. (2019). Classification of crop tolerance to heat and drought: A deep convolutional neural networks approach. *arXiv preprint arXiv:1906.00454*.

Khaki, S., and Wang, L. (2019). Crop yield prediction using deep neural networks. *Front. In Plant Sci.* 10, 621. doi: 10.3389/fpls.2019.00621

Khaki, S. (2019). *Source Code*. https://github.com/saeedkhaki92/CNN-RNN-Yield-Prediction.

Kim, N., Ha, K.-J., Park, N.-W., Cho, J., Hong, S., and Lee, Y.-W. (2019). A comparison between major artificial intelligence models for crop yield prediction: Case study of the midwestern united states, 2006–2015. *ISPRS Int. J. Geo-Information* 8, 240. doi: 10.3390/ijgi8050240

Kingma, D. P., and Ba, J. (2014). Adam: A method for stochastic optimization. *arXiv preprint arXiv:1412.6980*.

Kiranyaz, S., Ince, T., Abdeljaber, O., Avci, O., and Gabbouj, M. (2019). "1-d convolutional neural networks for signal processing applications," in ICASSP 2019-2019 IEEE International Conference on Acoustics, Speech and Signal Processing (ICASSP) (Brighton, United Kingdom: IEEE), 8360–8364.

LeCun, Y., Bengio, Y., and Hinton, G. (2015). Deep learning. *Nature* 521, 436. doi: 10.1038/nature14539

Liu, J., Goering, C., and Tian, L. (2001). A neural network for setting target corn yields. *Trans. ASAE* 44, 705. doi: 10.13031/2013.6097

Ng, A. Y. (2004). "Feature selection, L1 vs. L2 regularization, and rotational invariance," in *Proceedings of the Twenty-first International Conference on Machine learning* (Bnaff: ACM), 78.

Pham, V., Bluche, T., Kermorvant, C., and Louradour, J. (2014). "Dropout improves recurrent neural networks for handwriting recognition," in *14th International Conference on Frontiers in Handwriting Recognition* (Crete, Greece: IEEE), 285– 290.

Prasad, A. K., Chai, L., Singh, R. P., and Kafatos, M. (2006). Crop yield estimation model for iowa using remote sensing and surface parameters. *Int. J. Appl. Earth Observation Geoinformation* 8, 26–33. doi: 10.1016/j.jag.2005.06.002

Ransom, C. J., Kitchen, N. R., Camberato, J. J., Carter, P. R., Ferguson, R. B., Fernández, F. G., et al. (2019). Statistical and machine learning methods evaluated for incorporating soil and weather into corn nitrogen recommendations. *Comput. Electron. Agric.* 164, 104872. doi: 10.1016/j.compag.2019.104872

Ritchie, J. (1998). "Soil water balance and plant water stress," in *Understanding Options for Agricultural Production* (New York: Springer), 41–54.

Romero, J. R., Roncallo, P. F., Akkiraju, P. C., Ponzoni, I., Echenique, V. C., and Carballido, J. A. (2013). Using classification algorithms for predicting durum wheat yield in the province of buenos aires. *Comput. Electron. In Agric.* 96, 173–179. doi: 10.1016/j.compag.2013.05.006

Schauberger, B., Archontoulis, S., Arneth, A., Balkovic, J., Ciais, P., Deryng, D., et al. (2017). Consistent negative response of us crops to high temperatures in observations and crop models. *Nat. Commun.* 8, 13931. doi: 10.1038/ncomms13931

Shahhosseini, M., Martinez-Feria, R. A., Hu, G., and Archontoulis, S. V. (2019). Maize Yield and Nitrate Loss Prediction with Machine Learning Algorithms. *arXiv e-prints*, arXiv:1908.06746.

Sherstinsky, A. (2018). Fundamentals of recurrent neural network (rnn) and long short-term memory (lstm) network. *arXiv preprint arXiv:1808.03314*.

Sinclair, T., and Horie, T. (1989). Leaf nitrogen, photosynthesis, and crop radiation use efficiency: a review. *Crop Sci.* 29, 90–98. doi: 10.2135/cropsci1989.0011183x002900010023x

Springenberg, J. T., Dosovitskiy, A., Brox, T., and Riedmiller, M. (2014). Striving for simplicity: The all convolutional net. *arXiv preprint arXiv:1412.6806*.

Srivastava, N., Hinton, G., Krizhevsky, A., Sutskever, I., and Salakhutdinov, R. (2014). Dropout: A simple way to prevent neural networks from overfitting. *J. Mach. Learn. Res.* 15, 1929–1958.

Szegedy, C., Liu, W., Jia, Y., Sermanet, P., Reed, S., Anguelov, D., et al. (2015). Going deeper with convolutions (Cvpr). *2015 IEEE Conference on Computer Vision and Pattern Recognition,* Boston, MA.

Thornton, P., Thornton, M., Mayer, B., Wei, Y., Devarakonda, R., Vose, R. S. R., et al. (2018). *Daymet: Daily Surface Weather Data on a 1-km Grid for North America, Version 3*. (ORNL Distributed Active Archive Center). doi: 10.3334/ORNLDAAG/1328

Tibshirani, R. (1996). Regression shrinkage and selection *via* the lasso. *J. R. Stat. Soc. Ser. B (Methodological)* 58, 267–288. doi: 10.1111/j.2517-6161.1996.tb02080.x

USDA-NASS. (2019). *USDA - National Agricultural Statistics Service* Available at: https://www.nass.usda.gov/

Wang, A. X., Tran, C., Desai, N., Lobell, D., and Ermon, S. (2018). "Deep transfer learning for crop yield prediction with remote sensing data," in *Proceedings of the 1st ACM SIGCAS Conference on Computing and Sustainable Societies* (San Jose CA, USA: ACM), 50.

Yang, Q., Shi, L., Han, J., Zha, Y., and Zhu, P. (2019). Deep convolutional neural networks for rice grain yield estimation at the ripening stage using uav-based remotely sensed images. *Field Crops Res.* 235, 142–153. doi: 10.1016/j.fcr.2019.02.022

You, J., Li, X., Low, M., Lobell, D., and Ermon, S. (2017). "Deep gaussian process for crop yield prediction based on remote sensing data," in *Thirty-First AAAI Conference on Artificial Intelligence*, San Francisco, California, USA. 4559–4566.

**Conflict of Interest:** The authors declare that the research was conducted in the absence of any commercial or financial relationships that could be construed as a potential conflict of interest.